# A hierarchical framework for object recognition

Reza Moazzezi

Department of Electrical Engineering and Computer Sciences
University of California Berkeley

Email: rezamoazzezi@berkeley.edu

1-Abstract:

Object recognition in the presence of background clutter and distractors is a central problem both in neuroscience and in machine learning. However, the performance level of the models that are inspired by cortical mechanisms, including deep networks such as convolutional neural networks and deep belief networks, is shown to significantly decrease in the presence of noise and background objects [19, 24]. Here we develop a computational framework that is hierarchical, relies heavily on key properties of the visual cortex including mid-level feature selectivity in visual area V4 and Inferotemporal cortex (IT) [4, 9, 12, 18], high degrees of selectivity and invariance in IT [13, 17, 18] and the prior knowledge that is built into cortical circuits (such as the emergence of edge detector neurons in primary visual cortex before the onset of the visual experience) [1, 21], and addresses the problem of object recognition in the presence of background noise and distractors. Our approach is specifically designed to address large deformations, allows flexible communication between different layers of representation and learns highly selective filters from a small number of training examples.

2-Introduction

Neuroscience has played a key role in motivating a number of major machine learning algorithms. For example, deep networks [5], including convolutional neural networks [10] and HMAX model [16], are largely based on findings in primary visual cortex (V1); these models are inspired by the selectivity patterns of simple and complex cells in V1 and the repeating canonical structures observed throughout the whole cortex [23]. The main motivation behind our work is the failure of these cortically inspired models in the presence of background objects and distractors [19, 24]. Recently Szegedy et al. [24] performed a stability analysis on a range of deep networks including convolutional neural networks and showed that they are highly *unstable* in terms of recognition and a broad range of *imperceptible* perturbations can have a huge detrimental effect on their performance level. This is because in deep networks the classifiers develop selectivity to features that are different from the ones extracted by cortical networks. It is also likely that the feature extraction mechanisms prior to the classifier layer are not selective enough to ignore the perturbations (we discuss reference [19] in more detail in 'related work' section 5.1). In fact, the instabilities can appear as early as the first convolutional layer [24]. The problems reported in both cases [19, 24] are the result of the interference of a background noise that has never been presented within the training images. We refer to this as the "background interference" problem. For example, background interference problem arises when deep networks are trained on datasets where there is a single digit in each training image (such as MNIST dataset) and tested on datasets where there are multiple digits in each test image (figure 1 shows three example test images). We are not aware of any previous study that has provided a general solution for the background interference problem (For example, figures 1A, 1B and 1C show three example test images where there are multiple digits in each test image; in figure 1C, the background object overlaps the target). Hidden Markov Models (HMM) have been used on top of the convolutional neural networks to recognize sequences of digits (this is also referred to as the "sliding window" solution) [10]. However, since the sequential arrangement is necessary for the method suggested in [10], it does not offer a general solution. Another possible solution to background interference problem is to train the networks in the presence of background noise and distractors. There are major issues with this solution and we will discuss them in section 5.1.

The limitations discussed above are unlikely to arise from the feedforward structure of deep networks. Cortical processing is believed to be hierarchical [11]. In addition, the speed of processing in the human visual system is highly compatible with a feedforward architecture [25]. Therefore deep networks seem to miss a number of key cortical features that play central roles in neural information processing in the visual cortex. We propose that the sharp selectivity to mid-level features and the presence of prior structure in visual cortex are two possible key features that cortically inspired models should take into account. We briefly discuss them in this section (a more detailed discussion is provided in section 5).

Selectivity to features with moderate levels of complexity (mid-level features) is widely observed in Inferotemporal cortex and is believed to play a key role in object recognition [18] (this is discussed in detail in 'related work' section 5.4). In contrast, deep network models such as convolutional neural nets are designed such that they do not exhibit high degrees of selectivity. It is likely that the clutter tolerance observed in IT [14, 15, 31] is a result of its sharp selectivity to mid-level features.

In addition to sharp selectivity and clutter tolerance observed in IT, several studies have shown that a wide range of prior knowledge is already built into cortical circuits [1, 21, 27, 28]; edge detectors in the primary visual cortex are shown to emerge before the onset of the visual experience [1, 21] (this is discussed in detail in 'related work' section 5.3). An already built-in prior knowledge within the visual cortex might increase the speed of learning; in fact, Inferotemporal cortex (IT) develops its selectivity within 6 weeks after birth [27], which is extremely fast (We discuss further evidence that supports the presence of built-in prior knowledge in cortical circuits in 'related work' section 5.3).

In this work, we propose a hierarchical model that is heavily inspired by mid-level feature selectivity observed in IT and V4 [4, 9, 12, 13, 18] as well as the high degrees of selectivity and invariance observed in IT [13, 17, 18]. Our model learns from training datasets that contain only one object in each training image and successfully generalizes the learned features to test images with background objects (section 4.2 and 6.1). It also allows learning from very few examples and is fully translation and scale invariant (see 'results' section 4.1 and 'related work' section 5.2).

3-Computational Framework

3.1-Mid-level features

Inferotemporal cortex (IT) is one of the key brain regions involved in object recognition. Neurons in this area are selective to features that are more complex than edges but are less complex than objects. Such features are referred to as mid-level features. Ullman and colleagues [20] measured the mutual information between object categories and features with different levels of complexity and found that the mid-level features carry significantly more information than the features with higher or lower complexity levels. Therefore 1- mid-level complexity is the most informative level of complexity [20] and 2- higher level cortical circuits are selective to mid-level features [4, 9, 12, 13, 18]. In other words, both theory and experiments are pointing to the importance of the mid-level features. Mid-level features can also be defined as features that uniquely identify the objects. The results described in [20] show that such features carry moderate levels of complexity and therefore *the chances that mid-level features are generated randomly are very low due to their complexity level*. Therefore, if a mid-level feature is detected, it indicates the presence of the corresponding object class.

3.2-Recognition is invariant to a class of large perturbations (large perturbations principle)

Large deformations are a major source of variability within object classes and intra-class variability is a major problem for invariant recognition. We address this problem by introducing the concept of "minimal representation". An example digit 5 shown in figure 2 is approximated by a number of line segments (black lines). There are infinite ways to make such an approximation. However, we are interested in approximations where the number of the approximating lines is minimal; we refer to such an approximation as "minimal representation" (the approximation shown in figure 2 is a minimal representation). Note that the minimal representation is not unique (we describe how the minimal representation is constructed in methods section 7.1 and 7.2).

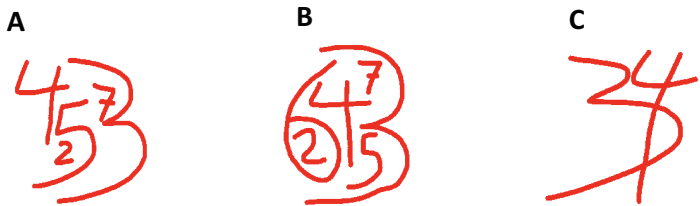

Figure 1) background interference problem. **A)** and **B)** example test images where multiple digits are presented simultaneously. **C)** target is a digit 4 and the background object (digit 3) overlaps the target. Deep networks such as convolutional neural networks that are trained on datasets that contain only one digit in each training image (such as MNIST dataset) fail to recognize them in the presence of visual clutter independent of whether the background objects overlap the target or not (see section 2).

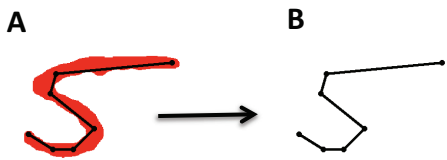

Figure 2) minimal representation. **A)** An example digit 5 and its minimal representation (**B**) (see methods sections 7.1 and 7.2).

Figure 3A (left) shows an example minimal representation of a digit 5. We refer to the corners of minimal representations as "key points". As mentioned above, minimal representation is not unique but any such representation has a property that we describe by the following example (figure 3): In figure 3A (right), we perturb one of the key points of the minimal representation (figure 3A (left)). The original angle of the key point is about 60 degrees (figure 3A (left)). After perturbation, the angle is about 120 degrees (figure 3A (right)). Despite this huge perturbation, human subjects still recognize it as digit 5. Figure 3B shows a large perturbation applied to the length of one of the edges. Figure 3C shows a perturbation where an edge is inserted between two consecutive edges of an arc (see figure 4 for the definition of arc). Finally, figure 3D shows several simultaneous perturbations. Human subjects recognize all these cases as digit 5 despite the large perturbations. Therefore *recognition seems to be invariant to a number of classes of large perturbations applied to the key points and the edges of minimal representations*. We refer to this as the "principle of large perturbations".

The principle of large perturbations can be used to generalize from a single training example. To this end, a range needs to be specified for the variables that can be perturbed (angles of the corners and lengths of the edges (including new edges added to the minimal representation)). For example, the range for angles can be (+/-) 60 degrees and the range for the lengths of the edges can be ½ to 2 times the original length. A range can also be defined for the lengths of the added edges (for example, at most twice the length of the longest neighboring edge).

The large perturbations principle can be used for recognition as well. We will discuss an efficient way to incorporate the large perturbations principle into our recognition model in section 3.3. Note that the minimal representation is simply a graph representation. It carries information about the connectivity structure of the vertices as well their geometry (x and y coordinates of the vertices). We will refer to the minimal representation as "simplified graph" throughout the text. In this framework, *a mid-level feature is a subgraph of the simplified graph*.

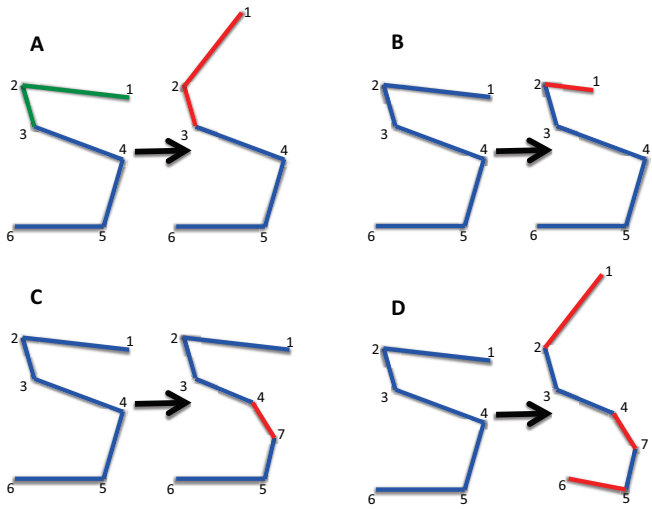

Figure 3) Recognition is invariant to large perturbations applied to the key points of the minimal representations. **A)** a large perturbation is applied to a key point.. **B)** a large perturbation is applied to the length of an edge. **C)** a new edge is added. **D)** several large perturbations are applied simultaneously.

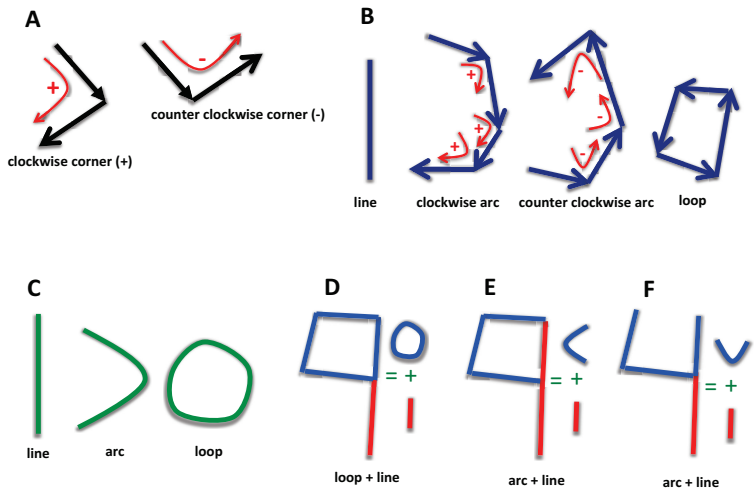

Figure 4) Decomposing digits into primitives. **A)** definition of clockwise and counter clockwise corners. **B)** definition of clockwise and counter clockwise arcs which is based on (A). A walk is a clockwise (counter clockwise) arc if all the corners are clockwise corners (counter clockwise corners) **C)** symbols used in this paper to represent lines, arcs and loops. **D)** decomposition of a mid-level feature that is extracted from the simplified graph of a digit 9 into loop and line. The whole simplified graph is assumed to be a mid-level feature. **E)** another decomposition of the simplified graph shown in (D) into an arc and a line. **F)** decomposition of a mid-level feature that is extracted from the simplified graph of a digit 4 into an arc and a line. Similar to (D) and (E), the whole simplified graph is assumed to be a mid-level feature.

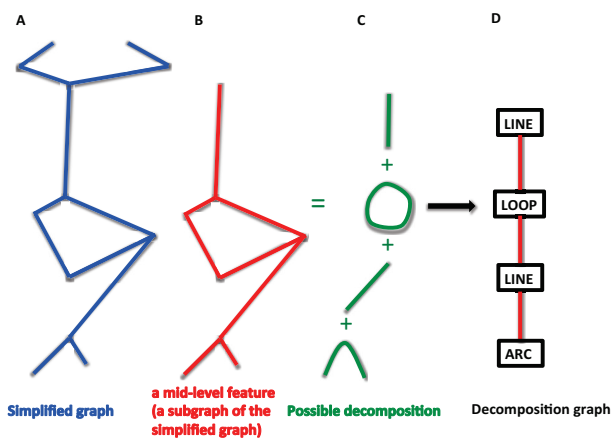

Figure 5) Extracting and decomposing a mid-level feature into primitives and building the decomposition graph. **A)** a cartoon of an example simplified graph. **B)** a subgraph of the simplified graph (a mid-level feature) shown in (A). **C)** a possible decomposition of the mid-level feature shown in (B). **D)** Decomposition graph. In the decomposition graph, the primitives are the nodes (represented by black rectangles) and the edges (red lines) represent the geometric relations. Note that the presence of an edge between two nodes in the decomposition graph shows that the properties of the geometric relation between the corresponding primitives is part of the representation of the corresponding mid-level feature class.

**3.3-Decomposition into primitives: an indirect way to apply large perturbations principle**

The large perturbations principle plays a key role in our model. It allows our generative model to generalize from single training examples [30]. One can think of this principle as a prior knowledge built into our object recognition model. The first step to incorporate this prior knowledge into our recognition model is to decompose the mid-level features (which as we discussed are subgraphs of the simplified graphs) into a number of primitives. In this paper, inspired by the selectivity patterns observed in IT, we consider three simple primitives: loops, arcs and straight lines (lines). (figures 4C-4F; their detailed definition is presented in the methods section)

1- arcs: an arc is a walk that has the following properties: walking along an arc, either all the turns are clockwise (towards right) or all the turns are counterclockwise (towards left). We refer to clockwise arcs as arc(+) and counter clockwise arcs as arc(-). The magnitude of the overall change of direction of an arc is less than or equal 360 degrees (see methods section for the definition of the overall change of direction; also see figure 6).
2- loops: a loop is defined as a closed walk within a graph.
3- straight lines: a straight line is a walk that has the following property: denoting the line segment connecting the two ends of a walk by $l$, if the maximum distance of the vertices of the walk from $l$ is less than a predefined threshold, then the walk is regarded as a straight line. Note that an arc that also satisfies the definition of a straight line can be regarded as both an arc and a straight line.

As an example, digit 9 in figure 4D can be decomposed into a loop and a straight line or digit 4 in figure 4F can be decomposed into an arc and a straight line. In both cases the whole simplified graphs are assumed to be a mid-level feature and therefore, in this case, decomposing the mid-level features into primitives is equivalent to decomposing the whole simplified graphs into primitives (in general, a mid-level feature can be any subgraph of the simplified graph). The digit 9 in figure 4D can also be decomposed into a straight line and an arc (figure 4E). In general having several decompositions improves the representational power of the model. The class specificity of mid-level features *orthogonalizes* the feature space and several ways of decomposition improves its robustness to noise.

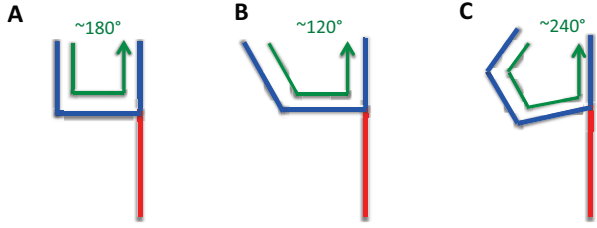

Figure 6) Mid-level feature classes. To generate a mid-level feature class, a mid-level feature is extracted from training examples, decomposed into primitives and a number of measurements are made on the primitives and their geometric relations. Next, a range is defined for each measurement. The mid-level feature class is defined by the decomposed graph, the measurements (made on the primitives and their geometric relations) and their corresponding ranges (see section 3.4 and section 7.7). Here we assume that **A**) is a mid-level feature extracted from a training example. It is decomposed into an arc (blue) and a straight line (red). A number of measurements are made on (A). One possible measurement is the overall change of direction of the arc. It is about 180 degrees for (A). Let's assume that the range defined by the mid-level feature class for this measurement (overall change of direction of the arc) is +/- 70 degrees (180 +/- 70). The mid-level features shown in **B**) and **C**) have the same decomposition as the mid-level feature shown in (A) and the overall change of direction for the arcs in both (B) and (C) are within the range defined for (A) (both 120 degrees (B) and 240 degrees (C) are between 180-70 degrees and 180+70 degrees). Similar to the overall change of direction of the arc, if all other measurements for (B) and (C) fall within the ranges defined for the corresponding measurements in (A), then (B) and (C) belong to the mid-level feature class defined by (A). Note that if we define another mid-level feature class on the mid-level feature shown in (A) where the range for the overall change of direction of the arc is +/- 50 degrees, then neither the mid-level feature shown in (B) nor the one shown in (C) belong to this new mid-level feature class (both 120 degrees (B) and 240 degrees (C) are not between 180-50 degrees and 180+50 degrees).

**3.4-Mid-level feature classes: definition, training, testing and measurements**

As we mentioned earlier, primitives provide us with an efficient way to apply the large perturbations principle during recognition: to this end, once a mid-level feature ($F$) is decomposed into primitives, a number of measurements are made on the primitives and their geometric relation, and a range is defined for each measurement (example measurements are the overall change of direction of an arc (see figure 6) or the direction of a straight line). In other words, instead of defining a range for the angles of the corners and the lengths of the edges, we define a range for the measurements made on the primitives and their geometric relation to form a mid-level feature class ($C_F$) defined on the mid-level feature $F$. Any other mid-level feature $F'$ belongs to the class $C_F$ if $F'$ has the same decomposition graph as $F$ (figure 5) and all the measurements on $F'$ are within the ranges specified by $C_F$. The mid-level feature class ($C_F$) is detected in an image if a mid-level feature that belongs to $C_F$ is detected in that image.

In detail, we define a number of *properties (measurements) for each primitive* and denote them by vector $\boldsymbol{P}^q = \{P_1^q, P_2^q, \ldots, P_{n_{(q)}}^q\}$ where $q$ is a primitive type (either a loop, arc or straight line; for example, $P_1^{loop}$ represents the length of the loops) and $n_{(q)}$ is the number of properties (measurements) defined on primitive $q$. We also define a number of *properties for the geometric relation between the primitives* and denote it by $\boldsymbol{P}^{\{q,r\}} = \{P_1^{\{q,r\}}, P_2^{\{q,r\}}, \ldots, P_{n_{\{q,r\}}}^{\{q,r\}}\}$ where $q$ and $r$ are primitive types and $n_{\{q,r\}}$ denotes the number of properties defined for the geometric relation between primitives $q$ and $r$ (for example, $P_1^{\{loop,arc\}}$ represents the direction of the vector connecting the centers of mass of a loop and an arc).

To construct a *mid-level feature class*, first, a mid-level feature $F$ is selected and decomposed into primitives (figure 5A-5C) and their properties are measured. Next, a number of pairs of primitives of $F$ are selected and the properties of their geometric relation are also measured. We denote the set of pairs of primitives of $F$ whose geometric relation is measured by $E$. We can represent the decomposed mid-level feature $F$ as a graph, $G$, where the vertices of $G$ represent the primitives and the edges of $G$ represent the geometric relation between the pairs of primitives that are specified by $E$. We refer to this representation, $G$, as the *decomposition graph* (figure 5D). We denote the set of the edges of $G$ by $E(G)$ and the set of the vertices of $G$ by $V(G)$.

We also refer to the primitive represented by a vertex in the decomposition graph as the type of that vertex and denote it with function $(T_{(.)})$; two vertices $j$ and $k$ are of the same type, that is, $T_j = T_k$, if they represent the same primitive type (for example if they both represent arc). Next we denote the properties of the primitive at vertex $j$ of the decomposition graph $G$ by $\boldsymbol{P}_j^{T_j} = \left\{P_{j,1}^{T_j}, P_{j,2}^{T_j}, \dots, P_{j,n_{(T_j)}}^{T_j}\right\}$ where $T_j$ represents the type of the primitive at vertex $j$ (or simply the type of vertex $j$). $n_{(T_j)}$ is the number of properties defined for the primitive at vertex $j$ (note that it only depends on its type $T_j$). We also denote the edge connecting vertices $j$ and $k$ by $\{j,k\}$. Similarly, we denote the properties of the geometric relation between the primitives at nodes $j$ and $k$ of $G$ by $\boldsymbol{P}_{\{j,k\}}^{\{T_j,T_k\}} = \left\{P_{\{j,k\},1}^{\{T_j,T_k\}}, P_{\{j,k\},2}^{\{T_j,T_k\}}, \dots, P_{\{j,k\},n_{\{T_j,T_k\}}}^{\{T_j,T_k\}}\right\}$ where $T_j$ and $T_k$ denote the type of the primitives at vertices $j$ and $k$, respectively. $\{T_j, T_k\}$ specifies the type of the geometric relation between vertices $j$ and $k$. $n_{\{T_j,T_k\}}$ is the number of properties defined for the geometric relation between primitives at vertices $j$ and $k$ (note that it only depends on their type $T_j$ and $T_k$). We represent the measurements on the primitives and their geometric relation by $\boldsymbol{M}_j^{T_j} = \left\{M_{j,1}^{T_j}, M_{j,2}^{T_j}, \dots, M_{j,n_{(T_j)}}^{T_j}\right\}$ and $\boldsymbol{M}_{\{j,k\}}^{\{T_j,T_k\}} = \left\{M_{\{j,k\},1}^{\{T_j,T_k\}}, M_{\{j,k\},2}^{\{T_j,T_k\}}, \dots, M_{\{j,k\},n_{\{T_j,T_k\}}}^{\{T_j,T_k\}}\right\}$, respectively ($M_{j,i}^{T_j}$ denotes the measurement on $P_{j,i}^{T_j}$ and $M_{\{j,k\},i}^{\{T_j,T_k\}}$ denotes the measurement on $P_{\{j,k\},i}^{\{T_j,T_k\}}$). We also define a range for each measurement (for both primitives and their geometric relation) and represent them by $\boldsymbol{R}_j^{T_j} = \left\{R_{j,1}^{T_j}, R_{j,2}^{T_j}, \dots, R_{j,n_{(T_j)}}^{T_j}\right\}$ and $\boldsymbol{R}_{\{j,k\}}^{\{T_j,T_k\}} = \left\{R_{\{j,k\},1}^{\{T_j,T_k\}}, R_{\{j,k\},2}^{\{T_j,T_k\}}, \dots, R_{\{j,k\},n_{\{T_j,T_k\}}}^{\{T_j,T_k\}}\right\}$ where $R_{j,i}^{T_j}$ is the range defined on measurement $M_{j,i}^{T_j}$ and $R_{\{j,k\},i}^{\{T_j,T_k\}}$ is the range defined on measurement $M_{\{j,k\},i}^{\{T_j,T_k\}}$.

Two decomposition graphs $G$ and $G'$ are **equivalent** if there exists a bijection between the set of vertices of $G$ and $G'$ ($f: V(G) \to V(G')$) such that for every adjacent pair of vertices $j$ and $k$ in $G$, $f(j)$ and $f(k)$ are adjacent in $G'$ (that is, if $\{j,k\} \epsilon E(G)$ then $\{f(j), f(k)\} \epsilon E(G')$) and $T_j = T_{f(j)}$ and $T_k = T_{f(k)}$.

**Definition of the mid-level feature class:**

A mid-level feature class associated with the mid-level feature $F$ and graph $G$ is defined by the quadruple $\{F, G, \boldsymbol{M}, \boldsymbol{R}\}$ where $\boldsymbol{M} = \left(\left\{\boldsymbol{M}_j^{T_j}\right\}_{\forall j \in V(G)}\right) \cup \left(\left\{\boldsymbol{M}_{\{j,k\}}^{\{T_j,T_k\}}\right\}_{\forall \{j,k\} \in E(G)}\right)$ and $\boldsymbol{R} = \left(\left\{\boldsymbol{R}_j^{T_j}\right\}_{\forall j \in V(G)}\right) \cup \left(\left\{\boldsymbol{R}_{\{j,k\}}^{\{T_j,T_k\}}\right\}_{\forall \{j,k\} \in E(G)}\right)$. In addition, any graph $G'$ that satisfies the following two conditions belongs to the mid-level feature class $\{F, G, \boldsymbol{M}, \boldsymbol{R}\}$:

**a)** $G'$ and $G$ are equivalent
**b)** if $\boldsymbol{N}_j^{T_j} = \left\{N_{j,1}^{T_j}, N_{j,2}^{T_j}, \dots, N_{j,n_{(T_j)}}^{T_j}\right\}$ and $\boldsymbol{N}_{\{j,k\}}^{\{T_j,T_k\}} = \left\{N_{\{j,k\},1}^{\{T_j,T_k\}}, N_{\{j,k\},2}^{\{T_j,T_k\}}, \dots, N_{\{j,k\},n_{\{T_j,T_k\}}}^{\{T_j,T_k\}}\right\}$ are the measurements on vertices and edges of $G'$, then the following two conditions hold:

$$M_{j,t}^{T_j} - R_{j,t}^{T_j} < N_{f(j),t}^{T_{f(j)}} < M_{j,t}^{T_j} + R_{j,t}^{T_j} \quad \forall j, t$$

$$M_{\{j,k\},t}^{\{T_j,T_k\}} - R_{\{j,k\},t}^{\{T_j,T_k\}} < N_{\{f(j),f(k)\},t}^{\{T_{f(j)},T_{f(k)}\}} < M_{\{j,k\},t}^{\{T_j,T_k\}} + R_{\{j,k\},t}^{\{T_j,T_k\}} \quad \forall \{j,k\} \in E(G), \forall t$$

We say a mid-level feature class is found in an image if a member of that class is found in that image. Note that the same decomposition graph $G$ can be combined with another set of ranges, $\boldsymbol{R}'$, to generate another mid-level feature class $\{F, G, \boldsymbol{M}, \boldsymbol{R}'\}$. Therefore it is possible to generate several mid-level feature classes from the very same mid-level feature. In fact, during training, for each mid-level feature $F$, a number of mid-level feature classes $\{F, G, \boldsymbol{M}, \boldsymbol{R}\}$, $\{F, G, \boldsymbol{M}, \boldsymbol{R}'\}, \{F, G, \boldsymbol{M}, \boldsymbol{R}''\}, \dots$ are generated and each of them are searched for independently in the validation set and the ones with *(nearly) zero false alarm rate* and *non-zero hit rate* are selected. Mid-level feature classes with (nearly) zero false alarm rate and non-zero hit rate are referred to as *highly informative mid-level feature classes* (the method used in this paper to search for the highly informative mid-level feature classes is discussed in detail in section 7.7). We call a mid-level feature $F$ highly informative if there exists a mid-level feature class $\{F, G, \boldsymbol{M}, \boldsymbol{R}\}$ that is highly informative.

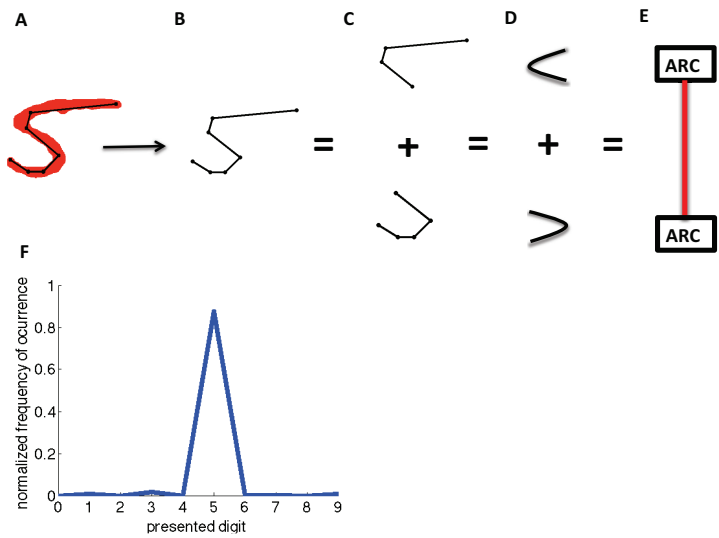

Figure 7) Highly informative mid-level feature class. **A)** An example digit 5 and its minimal representation (simplified graph). **B)** a mid-level feature is extracted from the simplified graph shown in (A). (The whole simplified graph is assumed to be a mid-level feature) **C)** and **D)** The mid-level feature shown in (B) is decomposed into two arcs. **E)** the decomposition graph associated with (C). **F)** the performance level of a highly informative mid-level feature class that is based on the decomposition graph shown in (E). It is detected in about 86% of digit 5s (about 86% hit rate) (F). It is rarely found in other digits (nearly zero false alarm rate). Therefore, the mid-level feature shown in (B) is a highly informative mid-level feature that uniquely identifies digit 5.

**Training:**

During the training phase, a number of mid-level features are selected and for each mid-level feature $F$, a number of mid-level feature classes $\{F, G, M, R\}$, $\{F, G, M, R'\}$, $\{F, G, M, R''\}$, … are generated. Each mid-level feature class is then searched for in different digits of the validation set and the highly informative mid-level feature classes (that have (nearly) zero false alarm rate and non-zero hit rate) are selected for the recognition phase (the method used to search for highly informative mid-level feature classes is described in detail in section 7.7).

**Testing:**

During recognition, our method searches for the highly informative mid-level feature classes (that were learned in the training phase) in test images. Since, by definition, highly informative mid-level feature classes have nearly zero false alarm rate, finding them in a test image is equivalent to recognition (for example, if a highly informative mid-level feature class is based on a mid-level feature that is extracted from a digit 3 in training dataset, then finding that highly informative mid-level feature class in a test image is equivalent to recognizing a digit 3 in that test image). Note that during recognition, it is possible that more than one highly informative mid-level feature class is found in a test image (for example, if there is more than one digit in the test image). This is discussed in detail in section 4.2.

**3.4.1-Measurements made on the primitives**

a) arcs: overall change of direction, direction of the line connecting the two ends of the arc
b) lines: direction of the line (which is defined by the direction of the line connecting the first vertex to the last vertex)

We did not define any measurements for the loops in this paper. The properties defined for the primitives (and their geometric relation) are very basic. As we will see in the results section, even such basic measurements result in a high performance level.

### 3.4.2-Measurements made on the geometric relation between the primitives

a) direction of the vector connecting their Centers of Mass
b) length of the vector connecting their centers of mass divided by the length of each primitive.
c) length of one of the primitives divided by the length of the other

Note that we only consider primitive-primitive relations here. Higher order relations can be broken into a number of second order relations (primitive-primitive relations).

The list of measurements made on decomposition graphs is not limited to the ones mentioned in section 3.4.1, 3.4.2 and the methods section. The exact nature of the measurements is subject to further research.

### 3.4.3-Coverage

We refer to the overall length of a mid-level feature divided by the overall length of the corresponding simplified graph as "coverage". Since in our framework, learning is based on single objects, coverage is well defined. It is a useful property when the objects, and therefore the resulting mid-level features, have simple structures (and the test image includes only one digit and the recognition system also knows that there is only one digit in the test image). In such cases, it is likely that the same mid-level feature is found in more than one object class. For example, the mid-level feature shown in figure 7D is found both in digit 5 (figure 7A) and in digit 3 (figure 9C) but its coverage is much higher for digit 5. Given the coverage, the mid-level feature shown in figure 7D has a very low chance of being detected in digit 3 and it uniquely identifies digit 5.

### 3.5-Advantages of mid-level feature class based representation

### 3.5.1-Mid-level feature classes: a combination of mid-level features, large perturbations principle and primitives

The mid-level feature class representation allows us to efficiently combine the idea of mid-level features with the principle of large perturbations. A major problem with the mid-level feature idea suggested by Ullman and colleagues [20] is the simple convolutional operation used to find the mid-level features in test images. We removed this limitation by decomposing the mid-level features into primitives and replacing the convolutional operator with large perturbations principle.

Furthermore, this representation allows our model to generalize from a single training example. More examples will help the model choose the best (highly informative) mid-level feature classes. In our experience, about 20 to 100 examples usually provide a very good estimate of the range for different measurements (see section 7.7 and figure 8).

### 3.5.2-Learning and recognition are fully automatic

In our model, we only specify the nature of the decompositions, primitives and their properties. Once these basic rules are specified, both learning and recognition are fully automatic. During learning, 1) each training example is turned into a minimal representation (simplified graph). 2) A number of subgraphs of the simplified graphs are automatically extracted, 3) decomposed into primitives and 4) turned into a number of mid-level feature classes. As discussed earlier, each subgraph can generate several mid-level feature classes. 5) Each mid-level feature class is searched for in all training examples and its hit rate and false alarm rate are measured. 6) Finally, highly informative mid-level feature classes are selected and used for recognition. Therefore the training pipeline is fully automatic.

During the test phase, the highly informative mid-level feature classes are searched for within each test image for classification. Since recognition is basically equivalent to graph search, recognition part of our method is also fully automatic.

### 3.6-An example illustrating our approach

The simplified graph of an example digit 5 is shown in figure 7A. Figure 7B shows an example mid-level feature extracted from the simplified graph shown in figure 7A. In this case, the whole simplified graph of the digit shown in figure 7A is assumed to be a mid-level feature $F$. One possible decomposition of $F$ results in two arcs (figure 7C and 7D). We refer to this decomposition as "arc + arc" decomposition. Figure 7E shows the decomposition graph ($G$). As mentioned earlier, a number of mid-level feature classes are generated for each mid-level feature. Figure 7F shows the performance of the best mid-level feature class (based on "arc + arc" decomposition) where the false alarm rate is nearly zero and the hit rate is about 86%. This mid-level feature class is learned based on 1000 training examples (100 examples per digit). Also note that training is performed on MNIST training dataset while testing is performed on a variant of MNIST test dataset where there is only one digit in each test image but, unlike MNIST test dataset, the test digits are at random locations and have different sizes (in MNIST dataset the digits are manually centered and resized to 28 by 28 pixels). As we will see in section 4.1, our model is fully translation and scale invariant.

In other words, figure 7 shows that there is a mid-level feature class based on "arc + arc" decomposition that only exists in digit 5 and accounts for 86% of digit 5s. This example shows a key difference between our approach and other cortically inspired models; in deep networks such as convolutional neural networks, even if an object is recognized correctly, it is not clear what feature(s) influenced the decision, nor it is clear where those feature(s) are extracted from (that is, where the object is). In contrast, in our model it is clear which feature was the basis of the decision and its location within the image is also immediately available (as mentioned earlier (section 3.2), in our model the recognition system is also aware of the absolute coordinates of the vertices of the simplified graphs).

Figure 8 shows the distribution of the measurements made on two properties of the mid-level feature class shown in figure 7. This includes the overall change of direction for the top arc and the $y$ component of the vector connecting the CoM of the two arcs ($\Delta COM_y$) divided by the length of the top arc ($L$) ($y$ axis is the vertical axis). As can be seen, 20 training examples give a good estimate of the range for these measurements.

The mid-level feature class in figure 7 has a very simple structure (the mid-level feature associated with this mid-level feature class has a decomposition structure of "arc + arc"). Digits are simple compared to real world objects and therefore the mid-level features that carry unique information about different digits normally turn out to be either the whole digits or a substantial part of them.

### 4-Experiments (results):

### 4.1-Informative mid-level features and recognition in the presence of location and scale variability

We trained our model on MNIST dataset where the digits are centered and resized to 28 by 28 pixels. During training, we randomly selected one or two mid-level features with moderate levels of complexity for each digit (see methods section 7.5 for details of the experiment; example highly informative mid-level features are shown in figure 9; selected mid-level features had high levels of coverage (coverage more than 90%); also note that as discussed in detail in the previous section, mid-level features are extracted from the training examples: a training example is selected and the corresponding simplified graph is extracted and a mid-level feature (a subgraph of the simplified graph) is extracted from the simplified graph). We defined a number of mid-level feature classes for each mid-level feature and measured their hit rate and false alarm rate on 100 examples per digit. We then selected the highly informative mid-level feature classes (described in detail in section 3.4; also see section 7.7 for details of how the range is selected for each measurement on the highly informative mid-level feature classes). Couple examples of highly informative mid-level features (alongside their decomposition graphs) are shown in figure 9.

We tested our model on a variant of MNIST dataset where the digits are no longer centered and scaled to 28 by 28 pixels (see methods section 7.5 for details of the experiment). The test digits are at random locations and their size is also random (note that in this case there is only one digit in each test image. The case where there are multiple digits in each test image is discussed in section 4.2). To measure the performance level of the learned highly informative mid-level feature classes, since there is only one digit in each image, we used coverage for classification: if two mid-level features are found within the same image, classification is based on the one that has a higher coverage. The overall performance level of the mid-level features for digit 2 (figures 9A and 9B) is about 85% while for digit 3 (figure 9C) is about 90%. The overall performance level of our model is about 85%. Note that we trained our model based on MNIST training dataset (where the digits are centered and resized to 28 by 28 pixels) but tested our model on a variant of MNIST where digits are at random locations and their size is also random. Since recognition in our model is based on graph representation, it is fully translation invariant. In addition, since relative lengths are measured (rather than absolute lengths) it is also fully scale invariant. Deep networks such as convolutional neural nets achieve 99% performance but, similar to the training dataset, the test dataset is also centered and resized to 28 by 28 pixels. We are not aware of any study that trains the deep networks on MNIST test dataset but tests them in the presence of location and scale variability.

Reference [26] showed that the performance level of deep belief networks (which is 99% for MNIST dataset) is 90% in the presence of very limited location variability (location variability of +/- 4 pixels within a 28 by 28 frame) and 94% in the presence of very limited size variability (20 by 20 to 28 by 28). They did not report the performance of their model for higher levels of variability. The performance level of our model is 85% and it is fully size and location invariant. To the best of our knowledge, other than [26], there is no systematic study to address the performance level of deep networks in the presence of location and size variability.

**4.2-Recognition in the presence of background objects (without overlap)**

We tested the performance level of the highly informative mid-level feature classes (learned from MNIST training dataset) on a test dataset that contained multiple digits in each test image. Digits in each test image had a significant variability in size and location and did not overlap (An example test image is shown in figure 10A. The experiments are described in detail in section 7.6). We found that the performance level of our method is exactly the same as the previous case (section 4.1) where there was only one digit in each test image (performance level is about 85%; performance is measured per digit). To understand this, note that once the (simultaneously presented) digits are turned into simplified graphs, since in this case the digits are separated from each other, their simplified graphs are also completely separated from each other (figure 10A shows both digits and their simplified graphs). Note that each highly informative mid-level feature class is independently searched for in all simplified graphs. In addition, each simplified graph is searched separately. Therefore, this is not different at all from the case in which there is only one digit in each test image (section 4.1). In addition, the performance level of our model is independent of the number of background objects; performance level remains at about 85% independent of the number background objects. Figure 10B shows the case where a novel background object (letter "M") is presented alongside the target (digit 4); again, our method recognizes the digit 4 and does not find any of the previously learned highly informative mid-level feature classes in letter "M".

As mentioned in section 3.2, the absolute coordinates of the simplified graphs are also measured; therefore when several digits are presented within an image, the absolute coordinates of all the simplified graphs are also extracted. This way, once a digit is recognized, our method has exact knowledge about its location. In addition, in our method, it is clear which highly informative mid-level feature class is the basis of the recognition. In contrast, in deep networks such as convolutional neural nets [10], even if they correctly recognize an object, neither the location of the object, nor the feature(s) that formed the decision are clear. As mentioned earlier, the case where the background objects overlap the target will be addressed in a separate paper. In section 6.1, we will present theoretical arguments that our method is also able to address the case where the digits overlap. This opens up the possibility of solving the background interference problem.

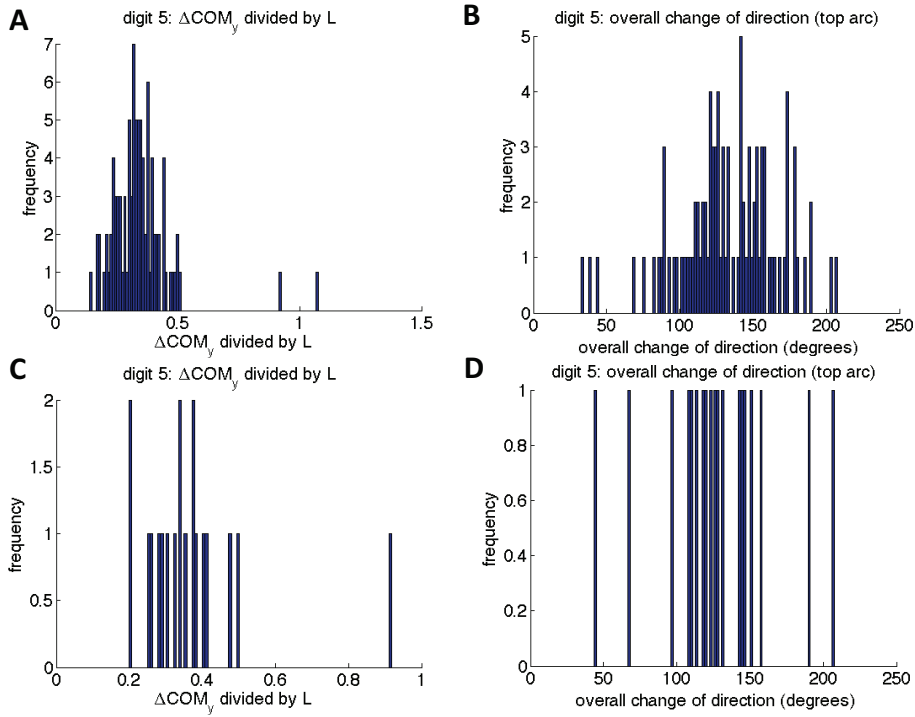

Figure 8) Small number of training examples provides a good estimate of the range. **A)** and **C)** Here we measure the projection on the $y$ axis (vertical axis) of the vector connecting the CoMs of the arcs shown in figure 7 divided by the length of the top arc. We measure this on 100 training examples (A) and 20 training examples (C). **B)** and **D)** The overall change of direction of the top arc in figure 7 is measured for 100 training examples (B) and 20 training examples (D). In both cases, 20 training examples gives a good estimate of the range for the measured properties.

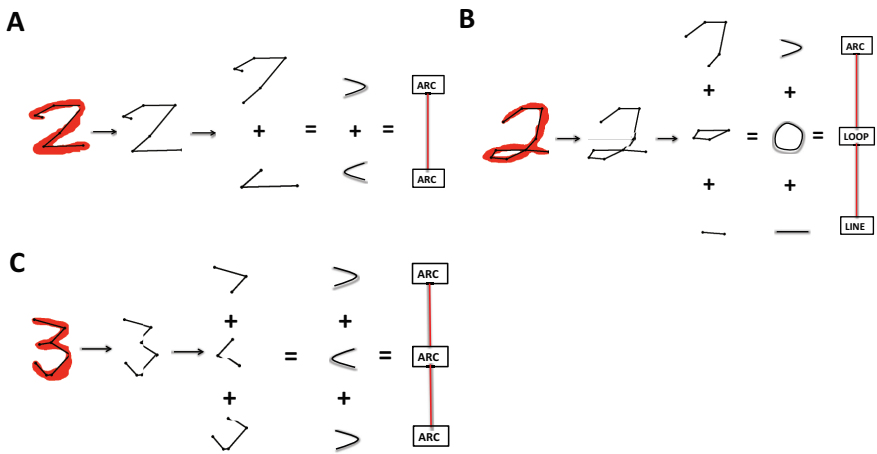

Figure 9) **A)** and **B)** two highly informative mid-level feature classes for digit 2. **C)** a highly informative mid-level feature class for digit 3. For each digit, the same conventions are used as in figure 7A to 7E. Note that in all cases, the whole simplified graphs are assumed to be a mid-level feature.

## 5-Related work

The hierarchical nature of the visual information processing in the brain is believed to play a central role in object recognition. In addition, inspired by Hubel and Wiesel's findings in primary visual cortex [6, 7], information processing in these models is implemented by an alternating sequence of linear and nonlinear operations [5, 16, 10]. Simple cells in primary visual cortex are modeled as "and" operation in HMAX and as "convolution" operation in convolutional neural networks while complex cells are modeled as "or" operation in HMAX and as "subsampling/pooling" operation in convolutional neural networks [10, 16]. This idea (also referred to as deep networks) has been examined extensively both in the field of neuroscience and machine learning and despite a number of encouraging results, a number of fundamental differences remain between these models and cortical processing. In section 2, we briefly discussed their difference in terms of developing selectivity to irrelevant features. We will discuss that in more detail in section 5.1. We will also discuss three other differences that are related to background interference problem in sections 5.2, 5.3 and 5.4: translation and scale invariance, built-in prior knowledge in cortical circuits and high degrees of selectivity in higher cortical areas.

### 5.1-Background interference problem

Similar to [24], [19] also reported that training Deep Belief Networks (DBN), another variant of deep networks, on MNIST training dataset and subsequently testing them on MNIST test dataset in the presence of a frame around each test digit results in about 60% drop in performance level despite the fact that the frames do not interact with (touch) the digits (note that the frames are only added to the test dataset). In other words, an uninformative frame affects the recognition and results in a huge drop in performance level. These results suggest that, similar to convolutional neural networks [24], DBNs develop selectivity to irrelevant features and get distracted by them. In contrast, high levels of clutter tolerance is observed in cortical networks [14, 15, 31]. It is not clear how the neural circuit develops highly selective neurons nor it is clear how high levels of clutter tolerance emerge within the neural system. It is likely that the presence of high degrees of selectivity and tolerance makes the visual system highly invariant to a broad range of noise and background objects.

One possible way for deep networks to address the background interference problem is to get trained in the presence of background objects. However, we are not aware of any study showing that successfully training deep networks in the presence of a limited number of background objects guarantees successful recognition in the presence of all other possible background objects. In the absence of such guarantees, it is likely that solving the background interference problem requires training deep networks on training datasets that grow exponentially with the number of background objects, their location, size, orientation, etc.

### 5.2-Translation and scale invariance

In deep networks, learning does not transfer from one location to another. In addition, translation invariant recognition of one object class does not transfer to other classes; this implies that the network, even if it achieves invariant recognition for the trained class, does not learn the transformation underlying the invariance and therefore is unable to generalize the transformation to other classes. Similarly, in deep networks, different scales need to be trained separately and scale invariance for one class does not transfer to other classes. In contrast, our method is fully scale and translation invariant.

One way to improve the performance level of deep networks is to train them at different locations, scales, etc; however, this results in the size of the training dataset to grow exponentially with the number of the training objects, their locations, sizes, etc. In addition, it is not clear if incorporating location and size variability within the training dataset necessarily improves the performance level. Sohn and Lee [26] added location and size variability to training dataset and showed that the performance level of DBNs, which is about 99% on MNIST dataset (where digits are manually centered and resized to 28 by 28 pixels), drops by up to 10% in the presence of limited variability in the location and size; adding location variability of (+/-) 4 pixels decreased the performance level by about 10%. They did not report the performance level for the case where the variability is only added to the test dataset. They only showed that adding variability to the training dataset does not improve the performance level to 99%. In other words, results provided in [26] shows that adding variability to training dataset does not necessarily improve the performance level.

**5.3-Prior knowledge is built into the cortical circuits**

The ultimate goal for any recognition model is to discover the highly informative features. However, studies discussed in section 2.1 show that the biologically inspired models such as deep networks fail to learn the key features. One way to fix this problem is to provide the object recognition system with some prior knowledge about the nature of the informative features. Such prior knowledge might help the network to learn the key features. For example, edge detectors in the primary visual cortex exist at the time of birth and are shown to be stable (by comparing the orientation maps at the time of birth and 6 month after birth; [1, 21]). Intracortical origins are proposed to explain the emergence of the edge detectors in the primary visual cortex [3]. In other words, the most basic type of selectivity in the cortex is already built into the cortical circuits and is not learned as a result of the brain's interactions with natural images. The existence of orientation selectivity in V1 at the time of birth can be interpreted as a prior knowledge given to cortical circuits to help them find the most informative features.

Furthermore, this does not seem to be limited to the primary visual cortex. A recent study [28] suggests that the extrastriate visual cortex in congenitally blind subjects responds to body-shape information suggesting that even the highest levels of object recognition pathway are developed independent of the visual experience. In [28] congenitally blind subjects were trained on body-shape information through auditory modality (by translating images into soundscapes). However, Striem-Amit and Amedi [28] found that it is the extrastriate visual cortex (and not the auditory cortex) that responds to the task relevant information. In addition, the area that responds to the task relevant information is the same area that responds to body-shape information in normal subjects (extrastriate body-selective area in visual cortex) [28]. These results suggest that the initial state of the visual cortex is not random and a wide range of prior knowledge is incorporated into the initial structure of the cortical circuits.

**5.4-Highly selective neurons in IT**

Neurons in IT area are highly selective to specific patterns with mid-level complexity [18]. While being highly selective, they also show high levels of invariance to location and size of the visual stimuli. In contrast, deep networks do not show such a sharp selectivity after training is complete [2]. The lack of high degrees of selectivity likely plays a key role in their sensitivity to irrelevant features [19, 24]. Note that HMAX model directly implements the "and" operation in "C" layers and therefore has the potential to exhibit high degrees of selectivity. However, it fails to account for high levels of variability observed in natural images [29]. In our model, high levels of selectivity is achieved through the highly informative mid-level feature classes. While the recognition is invariant to location, scale and a broad range of large perturbations, it is highly selective to decomposition graphs that form the highly informative mid-level feature classes.

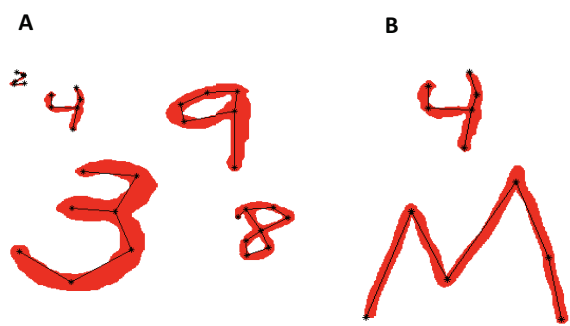

Figure 10) Recognition in the presence of background objects. **A)** a target digit (4) is presented together with a number of background digits (the simplified graphs are also shown (black lines)). **B)** a target digit (4) is presented in the presence of a novel background object (letter "M").

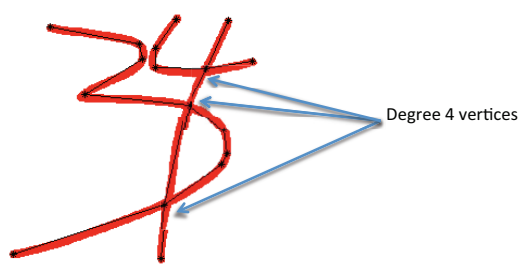

Figure 11) Highly overlapping digits. In this case, the overlap has not perturbed the simplified graphs associated with both digits. Degree 3 and 4 vertices are the sites of possible interactions and can expedite the search for the highly informative mid-level feature classes.

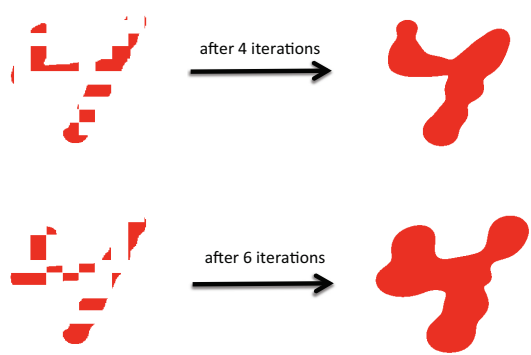

Figure 12) A recurrent network recovers the structure of a digit 4 with significant structural damage. The recovered structure can then be used for thinning and graph representation before the recognition stage. The filter used here was uniform. It is likely that V1 is the primary site for the neural implementation of this recurrent network.

## 6-Discussion

The performance level of deep networks on MNIST dataset is reported to be about 99%. However, this performance level is measured on a test dataset where there is only one digit in each test image and it is manually centered and resized to 28 by 28 pixels. The performance level of deep networks drops significantly if there are multiple digits in each test image (assuming that they are trained on MNIST training dataset where there is only one digit in each training image). We showed that the performance level of our method (that is trained on MNIST training dataset) is 85% and it is independent of the number of the background digits as long as they do not overlap the target. We will discuss the case where background objects overlap the target in section 6.1.

Our method is based on template matching but is different from conventional template matching techniques in two key ways: Template matching techniques have been criticized for 1- not being fully automatic and 2- for being computationally expensive. We have addressed both problems by 1- using mid-level feature-based techniques that makes both learning and recognition fully automatic (section 3.5.2) and 2- using the decomposition graph and the measurements made on the primitives and their geometric relations for template matching that significantly reduces the computational cost.

### 6.1-Overlapping digits

Our method is based on searching for highly informative mid-level feature classes within the simplified graphs of the digits. As discussed in the results section, if the simultaneously presented digits do not overlap, then the performance level of our model is independent of the number of background objects and is the same as the performance level for the case where there is a single object in each test image. As long as the highly informative mid-level feature classes are present within the simplified graph of the test images, an exhaustive search will find them.

When digits overlap, there is a possibility that each digit perturbs the simplified graph of the other digit. Figure 11 shows the example shown in figure 1C and its simplified graph. As can be seen, in this case the simplified graph of each digit is not perturbed by the other digit. In general, the interaction between simplified graphs is directly related to the process of thinning and simplification. In our experience, in most cases, overlapping digits do not perturb each other's simplified graphs (figure 11). If the simplified graphs are perturbed by the overlap, our simulations show that taking the width of the strokes into account removes the perturbations. We will present the simulations supporting this in a separate paper.

Once the problem of interaction between simplified graphs is addressed, the case where the digits overlap is not different from the case where they don't (section 4.2) and exhaustive search will find all the highly informative mid-level feature classes. This is because, since the overlap results in vertices with either degree 3 or higher, one can basically disentangle the digits at these vertices (figure 11). Note that disentangling the digits at higher degree nodes is just to demonstrate that the problem of overlapping digits is not different from the problem of non-overlapping digits as long as the overlap does not perturb the simplified graphs. Once the digits are separated from each other, the problem is basically equivalent to the problem of non-overlapping digits that we addressed in section 4.2. Also note that our model can take advantage of the vertices with higher order degrees to speed up the search process (if the computational power is limited).

### 6.2-Digits with structural damage

Figure 12 shows two digits with significant structural damage. A human subject recognizes them (provided that the subject knows that the stimulus is a digit) despite the major structural damages. The graph representations (simplified graphs) of such digits are noisy.

To address this problem, we rely on a mechanism similar to what we see in primary visual cortex. Recurrent interactions in V1 are believed to repeatedly process the input images through the same filter. The result of applying this operation on the structurally damaged digits is also shown in figure 12. The recurrent interactions gradually remove the structural damage and eventually recover the original digit. As can be seen, despite the substantial structural damage, the digits are recovered to a good extent after couple iterations. Since the number of layers needed for this process is normally large and that the selectivity of neurons is the same through the iterations (for figure 12, we used uniform filter), it is likely that the whole process is neurally implemented by recurrent interactions in primary visual cortex.

The recurrent network generates a new output at each iteration. In our method, the outputs of all iterations get simplified and the highly informative mid-level feature classes get searched for in all of them. For the digits shown in figure 12, the highly informative mid-level features are found after 4 and 6 iterations. Due to the complexity level of the highly informative mid-level features, it is unlikely that very basic operations, such as the recurrent interactions described above, generate a highly informative mid-level feature from noise. Therefore, the presence of highly informative mid-level features in the output of any of the iterations suggests the presence of the corresponding digits.

**6.3-Flexible interactions between different layers**

In our model, different layers of representation can easily communicate. For example, when digits overlap, the method can take advantage of the vertices with degree 3 or higher to speed up the search for highly informative mid-level feature classes. The vertices of the simplified graph are represented in lower layers of our hierarchical model while mid-level feature classes are represented in higher layers. Despite being in two different layers, they can send information back and forth to speed up the search process. Such a cross talk is very similar to feedback mechanisms in visual cortex that are missing in deep networks. Flexible communication between the fine and the coarse representations allows the graph-based representation to further explore the details of the images.

**6.4-Current limitations of the model**

We believe that thinning and simplification are two directions that can be improved. Our thinning/simplification method does not take the width of the strokes into account. Incorporating the width information is necessary to ensure that the edges of the simplified graphs closely follow the structure of the strokes. We are currently implementing this idea and the preliminary results are encouraging. As discussed earlier, it is also important to ensure that the simplified graphs of overlapping digits do not perturb each other.

Properties defined for primitives and their geometric relation is another direction that can be improved. There are several other properties that can be defined for primitives. For example, a loop can be a triangle or a square, etc. These measurements are not currently implemented but we can imagine datasets where such measurements can play an important role in recognition.

**6.5-Rotation invariant recognition**

In this work, we addressed the problem of location and size variability. We did not address rotation invariance. In the current implementation, the reference direction is positive direction on the $x$ axis. Removing the reference and defining the orientations relative to each other will make our model fully rotation invariant. This will allow us to learn from one orientation and generalize to all other orientations.

**6.6-Conclusion**

Extracting task relevant information while ignoring the task irrelevant information is the hallmark of the neural computation. In an object recognition task where a target object should be identified in an image, the target object is the task relevant information while its location, scale and the presence of other objects in the image are the task irrelevant information. Cortically inspired models, including deep networks such as convolutional neural networks [10], fail to generalize from one location or scale to another, which indicates that they fail to ignore this class of task irrelevant information. In addition, to the best of our knowledge, there is no study that systematically shows that they generalize the features that are learned from training datasets where there is a single object in each image to test datasets where there are multiple objects in each image (even if the objects do not overlap). In fact, there are reports that provide evidence against generalization of these models [19, 24]. This indicates that they also fail to ignore another class of task irrelevant information (presence of other objects in the image). In this work we suggest a computational framework that specifically addresses these problems. We find that our method is not only able to successfully generalize to images with background objects, but also learns from very few examples and is fully translation and scale invariant.

Our model relies on the key principles that govern the selectivity patterns of cortical neurons at both lower and higher layers of the visual cortex and offers a highly flexible solution for extracting the task relevant information in the presence of background objects and location and scale variability. Our method is able to learn from training datasets where there is only one digit in each training image and generalize the learned features to test datasets where there are multiple (non-overlapping) digits in each test image and opens up the possibility of achieving a human level performance on the problem of overlapping digits (where, again, learning is based on training datasets in which there is only one digit in each training image) and addressing the background interference problem in object recognition.

# 7-Methods

## 7.1-thinning

We tested our method on a variant of MNIST dataset where digits are at random locations and are of random sizes (either one digit in each test image or multiple digits with no overlap in each test image). To extract the minimal representation, we resize the original images (images are enlarged by a factor of 10) and apply a threshold to binarize them. For example, if the grayness level is between 0 and 1, any pixel above 0.1 is set to 1 (black pixel) and any pixel below 0.1 is set to zero (white pixel). We then apply the thinning algorithm developed in [22] to extract the skeleton of the binarized images. Resizing and thresholding are necessary to get the best results for the thinning algorithm.

## 7.2-simplification

Once a digit is thinned, we turn the thinned image into a graph (thinned graph) as follows: Each vertex of the thinned graph represents a black pixel in the thinned image. Let's denote the vertices of the thinned graph by $v_i$ and the corresponding pixels in the thinned image by $p_i$. Also let's denote the coordinates of pixel $p_i$ by $x_i$ and $y_i$. Two vertices $v_i$ and $v_j$ are directly connected by an edge in the thinned graph if:

1) $x_i = x_j$ and $y_i = y_j \pm 1$

2) $x_i = x_j \pm 1$ and $y_i = y_j$

3) $|x_i - x_j| = 1$ and $|y_i - y_j| = 1$ and there is no $v_k$ that satisfies the following two conditions:

    a) $|x_i - x_k| = 0$ and $|y_i - y_k| = 1$ and $|x_j - x_k| = 1$ and $|y_j - y_k| = 0$
    b) $|x_i - x_k| = 1$ and $|y_i - y_k| = 0$ and $|x_j - x_k| = 0$ and $|y_j - y_k| = 1$

Next, we simplify the thinned graph. There are several ways to simplify the thinned graph. Here we take the following approach: 1) we randomly choose two connected vertices of the thinned graph (let's denote the vertices by $v1$ and $v2$). We verify that all the vertices on the walk connecting them are of degree 2. 2) We connect the two vertices $v1$ and $v2$ with a line segment $l$. 3) If the line segment $l$ is completely within the digit (that is, if the line is completely covered by the black pixels), then we remove all the vertices between the two vertices $v1$ and $v2$ and directly connect $v1$ and $v2$ by a new edge. We repeat steps 1 to 3 until any new connecting line segment is not completely covered by black pixels. The output of this process is the simplified graph (minimal representation).

## 7.3-decomposition into primitives

Once a simplified graph is extracted, a number of subgraphs are selected to form a number of mid-level feature classes. Each subgraph is then decomposed into loops, arcs and lines. Ideally, all subgraphs are extracted and decomposed. However, if the computational power is limited, then the priority is given to subgraphs with moderate levels of complexity [20]. Here we define the primitives, arcs, loops and straight lines:

**arcs:**

An "arc" is a walk $V_0 \to E_0 \to V_1 \to E_1 \to V_2 \to E_2 \to \cdots \to E_n \to V_{n+1}$ in the graph that does not visit the same vertex twice ($V_i$ denotes the $i$th vertex and $E_i$ denotes the $i$th edge of the walk). In addition, moving from one end of the arc towards the other, the turns are either always clockwise (we refer to such an arc as "arc+") or counter clockwise ("arc-"). **d)** We define the length of the arc as the sum of the length of the edges making the arc ($L_{arc} = \sum_i l_{E_i}$) where $l_{E_i}$ denotes the length of the edge $E_i$.

**straight lines:**

A "straight line" is a walk in the graph that does not visit the same vertex twice. In addition, the maximum deviation of the nodes on the walk from the line connecting the two ends of the walk is less than a threshold (we will use "line" and "straight line" interchangeably): Let's $V_0 \to E_0 \to V_1 \to E_1 \to V_2 \to E_2 \to \cdots \to E_n \to V_{n+1}$ represent a walk where $V_i$ is the $i$th vertex and $E_i$ is the $i$th edge of the walk. We denote the vector that connects vertex 0 to vertex $(n+1)$ by $\overrightarrow{D_{line}}$. We denote the distance between vertex $j$ and the vector $\overrightarrow{D_{line}}$ by $d_{j,\overrightarrow{D_{line}}}$. For a straight line we have:

$$max_j \left(\frac{d_{j,\overrightarrow{D_{line}}}}{L_{line}}\right) < threshold$$

where $L_{line}$ is the length of the line ($L_{line} = \sum_i l_{E_i}$ where $l_{E_i}$ is the length of the edge $E_i$).

**loops:**

A loop is a closed walk, $V_0 \rightarrow E_0 \rightarrow V_1 \rightarrow E_1 \rightarrow V_2 \rightarrow E_2 \rightarrow \cdots \rightarrow E_n \rightarrow V_0$. The length of the loop is defined as ($L_{arc} = \sum_i l_{E_i}$) where $l_{E_i}$ denotes the length of the edge $E_i$.

**7.4-measurements made on the primitives and their geometric relation**

**arcs:**

**a)** overall change of direction ($\Delta\theta$): Consider an arc $V_0 \rightarrow E_0 \rightarrow V_1 \rightarrow E_1 \rightarrow V_2 \rightarrow E_2 \rightarrow \cdots \rightarrow E_n \rightarrow V_{n+1}$ where $V_i$ is the $i$th vertex and $E_i$ is the $i$th edge of the arc. We denote the change of direction at vertex ($i$) (that is, as you move from edge ($i$-1) to edge ($i$)) by $\Delta\theta_i$. Since all the turns in an arc are either always clockwise or always counter clockwise, all $\Delta\theta_i$s are either always positive or always negative. The magnitude of each $\Delta\theta_i$ is between 0 and 180 degrees. We define the overall change of direction of an arc as follows:

$$\Delta\theta = \sum_i \Delta\theta_i$$

**b)** direction of arcs ($\overrightarrow{D_{arc}}$): We denote the vector connecting vertex 0 to vertex ($n+1$) by $\overrightarrow{D_{arc}}$. We define the direction of the arc as the direction of $\overrightarrow{D_{arc}}$.

**c)** arc-normal ($\overrightarrow{N_{arc}}$): $\overrightarrow{N_{arc}}$ is a vector orthogonal to $\overrightarrow{D_{arc}}$ and its direction is away from the arc.

**d)** direction of $E_0$ and $E_n$

**straight lines:**

**a)** direction of the line ($\overrightarrow{D_{line}}$): consider a straight line $V_0 \rightarrow E_0 \rightarrow V_1 \rightarrow E_1 \rightarrow V_2 \rightarrow E_2 \rightarrow \cdots \rightarrow E_n \rightarrow V_{n+1}$. We denote the vector that connects vertex 0 to vertex ($n+1$) by $\overrightarrow{D_{line}}$. We define the direction of the line as the direction of the vector $\overrightarrow{D_{line}}$.

**geometric relations:**

The properties defined for pairs of primitives are as follows:

**a)** relative length of the primitives ($L_1/L_2$) where $L_1$ and $L_2$ are the length of the two primitives.

**b)** vector connecting the centers of mass of the two primitives ($\overrightarrow{\delta cm}$, center of mass is defined as the center of mass of the vertices of the primitives) normalized by the length of the primitives ($\overrightarrow{\delta cm}/L_1, \overrightarrow{\delta cm}/L_2$) as well as normalized by the length of the whole graph ($\overrightarrow{\delta cm}/L_G$) (We assume that there is only one digit in each training image and therefore the length of the whole graph is well-defined).

**c)** relative overall change of directions (if both primitives are arcs) ($CD_1/CD_2$): where $CD_1$ and $CD_2$ are the overall change of directions of the two arcs.

**d)** connection point (for loops): (when one primitive is loop and the other one is either an arc or a line) loops are divided into four quarters and this property specifies at which quarter the line or arc are connected to the loop. We can also connect the two ends of an arc and define the connection point for arcs as well.

**e)** relative directions of lines: Whether the difference of the directions of two lines are positive or negative as well as the magnitude of the difference. This can also be extended to arcs. The direction of an arc can also be compared to the direction of another arc or another line.

**7.5-experiments: single digit in each test image**

The size of the MNIST images are 28 by 28. We resized the images (both training and test dataset) to 280 by 280 (using matlab imresize function) and then binarized them (pixel values more than 0.1 were set to 1 otherwise 0). We used these images for thinning and simplification. As mentioned earlier, resizing and thresholding are necessary to get the best results for the thinning method.

To test our model, in particular in the presence of size and location variation, we created a test dataset as follows. Each test image was resized to 280 by 280 pixels and then binarized. The resulting images were then randomly rescaled (the range between 84 by 84 to 560 by 560), and randomly located within a larger image (1000 by 1000). Note that the range tested in simulations is limited because of the limited computational power. As described in previous sections, our method is fully scale and translation invariant. Note that in this case, there is only one digit in each test image (the case where there are multiple digits in each test image is discussed in section 4.2 and 7.6).

We randomly selected one or two mid-level features (with coverage more than 90%) for each digit. If a digit can be represented by more than one decomposition graph (for example figures 9A and 9B) we made sure that there is a mid-level feature associated with each of them (and therefore for some digits we extracted more than one mid-level feature).

To measure the performance level of the learned highly informative mid-level feature classes, since there is only one digit in each test image, we used the coverage for classification: if two mid-level features are found within the same image, classification is based on the one that has a higher coverage.

**7.6- experiments: multiple digits in each test image (non-overlapping case)**

To generate the test images for this case, the 280 by 280 pixel MNIST test digits generated in section 7.5 were randomly resized (between 81 by 81 to 1000 by 1000) and randomly located within an image of size 2000 by 2000 pixels. The image was then binarized (see section 7.5). We made sure that the digits did not overlap in this case.

**7.7-finding the range for each measurement made on highly informative mid-level feature classes**

In the main text we discuss defining a number of mid-level feature classes for each mid-level feature where different classes differ only in their set of ranges (that are defined on the measurements). In detail, the process to find the highly informative mid-level feature classes is as follows: once a mid-level feature $F$ is extracted and decomposed into primitives ($D$), one can search for $D$ in a number of training examples. We index the properties defined on $D$ (including properties for the primitives and their geometric relation) by $i$. We denote the measurement made on property $i$ on training example $j$ by $m_{i,j}$. For measurement $i$, one can then estimate the range by $(\min_j(m_{i,j}), \max_j(m_{i,j}))$. This is the estimated range for measurement $i$ based on the training examples. In general, the complexity level of mid-level features makes it unlikely that two highly informative mid-level features share the same decomposition graph. In our experience, the range $(\min_j(m_{i,j}), \max_j(m_{i,j}))$ for all measurements $i$ establishes the range for highly informative mid-level feature classes for large number of training examples. However, for small number of training examples (see figure 8), it is helpful to also define mid-level feature classes with slightly larger range $(\min_j(m_{i,j}) - a_i, \max_j(m_{i,j}) + b_i)$ for each measurement $i$ where $a_i$s and $b_i$s are positive numbers. For small number of training examples, it is likely that the mid-level feature class with slightly larger range for its measurements performs better on the validation set. This is simply because the number of training examples is small and therefore the range learned from the training examples does not include other examples whose measurements might fall outside of the range. Note that for small number of training examples, one can define several candidate mid-level feature classes including the one whose range is based on the training examples, $(\min_j(m_{i,j}), \max_j(m_{i,j}))$ for all measurements $i$, and those whose ranges are slightly larger $(\min_j(m_{i,j}) - a_i, \max_j(m_{i,j}) + b_i)$ for all measurements $i$ and measure their hit rate and false alarm rate on validation set and select the one with highest hit rate (and (nearly) zero false alarm rate). Therefore, to find the highly informative mid-level feature classes for small number of training examples, we start from the ranges that are based on the measurements that are made on the training examples.

**References:**

[1] Crair, M. C., Gillespie, D. C., & Stryker, M. P. (1998). The role of visual experience in the development of columns in cat visual cortex. *Science*, *279*(5350), 566-570.

[2] Erhan, D., Bengio, Y., Courville, A., & Vincent, P. (2009). Visualizing higher-layer features of a deep network. *Dept. IRO, Université de Montréal, Tech. Rep*.

[3] Ernst, U. A., Pawelzik, K. R., Sahar-Pikielny, C., & Tsodyks, M. V. (2001). Intracortical origin of visual maps. *nature neuroscience*, *4*(4), 431-436.

[4] Kobatake, E., & Tanaka, K. (1994). Neuronal selectivities to complex object features in the ventral visual pathway of the macaque cerebral cortex. *Journal of neurophysiology*, *71*, 856-856.

[5] Fukushima, K. (1988). Neocognitron: A hierarchical neural network capable of visual pattern recognition. *Neural networks*, *1*(2), 119-130.

[6] Hubel, D. H., & Wiesel, T. N. (1959). Receptive fields of single neurones in the cat's striate cortex. *The Journal of physiology*, *148*(3), 574.

[7] Hubel, D. H., & Wiesel, T. N. (1962). Receptive fields, binocular interaction and functional architecture in the cat's visual cortex. *The Journal of physiology*, *160*(1), 106.

[8] Kavukcuoglu, K., Ranzato, M., Fergus, R., & LeCun, Y. (2009, June). Learning invariant features through topographic filter maps. In *Computer Vision and Pattern Recognition, 2009. CVPR 2009. IEEE Conference on* (pp. 1605-1612). IEEE.

[9] Gallant, J. L., Braun, J., & Van Essen, D. C. (1993). Selectivity for polar, hyperbolic, and Cartesian gratings in macaque visual cortex. *Science*, *259*(5091), 100-103.

[10] LeCun, Y., Bottou, L., Bengio, Y., & Haffner, P. (1998). Gradient-based learning applied to document recognition. *Proceedings of the IEEE*, *86*(11), 2278-2324.

[11] Maunsell, J. H., & Newsome, W. T. (1987). Visual processing in monkey extrastriate cortex. *Annual review of neuroscience*, *10*(1), 363-401.

[12] Pasupathy, A., & Connor, C. E. (1999). Responses to contour features in macaque area V4. *Journal of Neurophysiology*, *82*(5), 2490-2502.

[13] Kobatake, E., & Tanaka, K. (1994). Neuronal selectivities to complex object features in the ventral visual pathway of the macaque cerebral cortex. *Journal of neurophysiology*, *71*, 856-856.

[14] Zoccolan, D., Kouh, M., Poggio, T., & DiCarlo, J. J. (2007). Trade-off between object selectivity and tolerance in monkey inferotemporal cortex. *The Journal of Neuroscience*, *27*(45), 12292-12307.

[15] Chelazzi, L., Duncan, J., Miller, E. K., & Desimone, R. (1998). Responses of neurons in inferior temporal cortex during memory-guided visual search. *Journal of Neurophysiology*, *80*(6), 2918-2940.

[16] Riesenhuber, M., & Poggio, T. (2000). Models of object recognition. *Nature neuroscience*, *3*, 1199-1204.

[17] Tanaka, K., Saito, H. A., Fukada, Y., & Moriya, M. (1991). Coding visual images of objects in the inferotemporal cortex of the macaque monkey. *J Neurophysiol*, *66*(1), 170-189.

[18] Tanaka, K. (1996). Inferotemporal cortex and object vision. *Annual review of neuroscience*, *19*(1), 109-139.

[19] Tang, Y., & Eliasmith, C. (2010). Deep networks for robust visual recognition. In*Proceedings of the 27th International Conference on Machine Learning (ICML-10)* (pp. 1055-1062).

[20] Ullman, S., Vidal-Naquet, M., & Sali, E. (2002). Visual features of intermediate complexity and their use in classification. *Nature neuroscience*, *5*(7), 682-687.

[21] Wiesel, T. N., & Hubel, D. H. (1974). Ordered arrangement of orientation columns in monkeys lacking visual experience. *Journal of comparative neurology*, *158*(3), 307-318.

[22] Wu, R. Y., & Tsai, W. H. (1992). A new one-pass parallel thinning algorithm for binary images. *Pattern Recognition Letters*, *13*(10), 715-723.

[23] Douglas, R. J., Martin, K. A., & Whitteridge, D. (1989). A canonical microcircuit for neocortex. *Neural computation*, *1*(4), 480-488.

[24] Szegedy, C., Zaremba, W., Sutskever, I., Bruna, J., Erhan, D., Goodfellow, I., & Fergus, R. (2013). Intriguing properties of neural networks. *arXiv preprint arXiv:1312.6199*.

[25] Thorpe, S., Fize, D., & Marlot, C. (1996). Speed of processing in the human visual system. *nature*, *381*(6582), 520-522.

[26] Sohn, K., & Lee, H. (2012). Learning invariant representations with local transformations. *arXiv preprint arXiv:1206.6418*.

[27] Rodman, H. R., Scalaidhe, S. P. O., & Gross, C. G. (1993). Response properties of neurons in temporal cortical visual areas of infant monkeys.*Journal of Neurophysiology*, *70*, 1115-1115.

[28] Striem-Amit, E., & Amedi, A. (2014). Visual Cortex Extrastriate Body-Selective Area Activation in Congenitally Blind People "Seeing" by Using Sounds. *Current Biology*, *24*(6), 687-692.

[29] Pinto, N., Cox, D. D., & DiCarlo, J. J. (2008). Why is real-world visual object recognition hard?. *PLoS computational biology*, *4*(1), e27.

[30] Miller, E. G., Matsakis, N. E., & Viola, P. A. (2000). Learning from one example through shared densities on transforms. In *Computer Vision and Pattern Recognition, 2000. Proceedings. IEEE Conference on* (Vol. 1, pp. 464-471). IEEE.

[31] Reynolds, J. H., Chelazzi, L., & Desimone, R. (1999). Competitive mechanisms subserve attention in macaque areas V2 and V4. *The Journal of Neuroscience*,*19*(5), 1736-1753.